\documentclass[12pt]{article}

\usepackage{graphicx}
\usepackage{caption}
\usepackage{subcaption}
\usepackage{mathptmx}           
\usepackage[applemac]{inputenc} 
\usepackage{clrscode}             
\usepackage{url} 
\usepackage{color}
\usepackage[usenames,dvipsnames]{xcolor}
\usepackage{ marvosym }
\usepackage{enumitem}
\usepackage{natbib}
\usepackage{hyperref}

\begin{document}
\sloppy

\title{\bf A Java Implementation of the SGA, UMDA, ECGA, and HBOA}

\author{   {\bf José C. Pereira}\\
            \small CENSE and DEEI-FCT\\
            \small Universidade do Algarve\\
            \small Campus de Gambelas\\
            \small 8005-139 Faro, Portugal\\
            \small unidadeimaginaria@gmail.com
\and
		 {\bf Fernando G. Lobo}\\
            \small CENSE and DEEI-FCT\\
            \small Universidade do Algarve\\
            \small Campus de Gambelas\\
            \small 8005-139 Faro, Portugal\\
            \small fernando.lobo@gmail.com         
}
\date{}
\maketitle

\begin{abstract}

The Simple Genetic Algorithm, the Univariate Marginal Distribution Algorithm, the Extended Compact Genetic Algorithm, and the Hierarchical Bayesian Optimization Algorithm are all well known Evolutionary Algorithms. 

In this report we present a Java implementation of these four algorithms with detailed instructions on how to use each of them to solve a given set of optimization problems. Additionally, it is explained how to implement and integrate new problems within the provided set. The source and binary files of the Java implementations are available for free download at \textcolor{Blue}{\href{https://github.com/JoseCPereira/2015EvolutionaryAlgorithmsJava}{https://github.com/JoseCPereira/2015EvolutionaryAlgorithmsJava}}. 
 
\end{abstract}


\section{Introduction}\label{sec:intro}

Evolutionary Algorithms (EAs) are a group of optimization techniques inspired by computational models of evolutionary processes from biology, namely natural selection and survival of the fittest. The idea of using biological evolution as a paradigm for stochastic problem solvers has been around since the 1950s and it experienced major developments in the 1970s and 1980s with the emergence of four main independent research fields: Genetic Algorithms (GA), Evolutionary Strategies (ES), Evolutionary Programming (EP) and Genetic Programming (GP). Over the last two decades new classes of Evolutionary Algorithms have been developed, one of them being Estimation of Distribution Algorithms (EDAs). Currently, EAs are a very active and dynamic research field, with various applications in many areas of scientific and technological knowledge. 

In this report we present Java implementations of the Simple Genetic Algorithm (SGA) \citep{Holland:75,Goldberg:89a,EibenSmith:03}, and of three well known EDAs: the Univariate Marginal Distribution Algorithm (UMDA) \citep{Muhlenbein:96}, the Extended Compact Genetic Algorithm (ECGA) \citep{Harik:99a}, and the Hierarchical Bayesian Optimization Algorithm (HBOA) \citep{PelikanGoldberg:06}. Each of these algorithms is a well known EA with a significant amount of published research. We encourage the interested reader to explore the existing rich scientific literature on the subject.

The Java implementations of these four EAs follow the same basic design, with similar I/O interface and parameter configuration.
All implementations satisfy the same set of constraints:

\begin{enumerate}
\item The algorithm represents possible solutions (individuals) as strings of zeros and ones.
\item All individuals have the same string size.
\item The population size remains constant throughout a complete run of the algorithm.
\end{enumerate}

In the following, we provide detailed instructions on how to use each of the algorithms to solve optimization problems. A set of well known optimization problems is given and, additionally, it is explained how to implement and integrate new problems within the provided set. The source and binary files of the Java implementations are available for free download at \textcolor{Blue}{\href{https://github.com/JoseCPereira/2015EvolutionaryAlgorithmsJava}{https://github.com/JoseCPereira/2015EvolutionaryAlgorithmsJava}}.

The implemented set of standard EAs presented in this report is also the basis for the Java implementation of the \emph{Parameter-less Evolutionary Algorithms} as discussed in another arXiv report and whose corresponding source is also available for free download at \textcolor{Blue}{\href{https://github.com/JoseCPereira/2015ParameterlessEvolutionaryAlgorithmsJava}{https://github.com/JoseCPereira/2015ParameterlessEvolutionaryAlgorithmsJava}}.

\section{How to use the code}\label{sec:howto}

Herein we provide detailed instructions on how to use a Java implementation of the generic algorithm \emph{EA}, where the acronym is a mere place holder for the algorithms SGA, UMDA, ECGA, and HBOA.
 
The EA is a Java application developed with the Eclipse\footnote{Version: Kepler Service Release 2} IDE. The available code is already compiled and can be executed using the command line.

 \paragraph{Run the EA from a command line}
 
\begin{enumerate}
\item Unzip the source file \textit{2015EAs.zip} to any directory.
\item Open your favourite terminal and execute the command\\ 
\vspace{-.5cm}

{\tt{~~cd [yourDirectory]/2015EvolutionaryAlgorithmsJava/EA/bin}}\\
\vspace{-.3cm}

where {\tt{[yourDirectory]}} is the name of the directory chosen in step 1.
 
\item Execute the command\\ 
\vspace{-.5cm}

{\tt{~~java com/EA/EA EAParameters.txt}}
\end{enumerate}

\vspace{.3cm}
The argument ``EAParameters.txt'' is in fact the name of the file containing all the options concerning the parameter settings and can be changed at will.

After each execution of a single or multiple runs, the EA produces one output file -- \emph{EA\_*\_*.txt} -- that records how each run progressed in terms of fitness calls, best  current fitness, among other relevant information. Additionally, the EA also creates the file -- \emph{EA-STATS\_*\_*.txt} -- that stores some of the statistics  necessary for analyzing the behaviour of the EA over multiple runs.

 \subsection{Set of optimization problems}
 
The current code includes a set of optimization problems that can be used to test any of the four EAs. Here is the problem menu:\\

\textit{ZERO} Problems \textit{\qquad \qquad \qquad \qquad \qquad ONE} Problems
\vspace{.4cm} 

\begin{tabular}{rlrl}
~~0 $\rightarrow$ & ZeroMax & ~~10 $\rightarrow$ & OneMax\\
~~1 $\rightarrow$ & Zero Quadratic & ~~11 $\rightarrow$ & Quadratic\\
~~2 $\rightarrow$ & Zero 3-Deceptive & ~~12 $\rightarrow$ & 3-Deceptive\\
~~3 $\rightarrow$ & Zero 3-Deceptive Bipolar & ~~13 $\rightarrow$ & 3-Deceptive Bipolar\\
~~4 $\rightarrow$ &  Zero 3-Deceptive Overlapping & ~~14 $\rightarrow$ & 3-Deceptive Overlapping\\
~~5 $\rightarrow$ & Zero Concatenated Trap-k & ~~15 $\rightarrow$ & Concatenated Trap-k\\
~~6 $\rightarrow$ & Zero Uniform 6-Blocks & ~~16 $\rightarrow$ & Uniform 6-Blocks
\end{tabular}\\ 

\

\textit{Hierarchical} Problems	
\vspace{.4cm} 

\begin{tabular}{rl}
~~21 $\rightarrow$ &  Hierarchical Trap	One\\
~~22 $\rightarrow$ & Hierarchical Trap	Two
\end{tabular}\\

\

The Zero problems always have the string with all zeros as their best individual. The One problems are the same as the Zero problems but their best individual is now the string with all ones. A description of these problems can be found, for instance, in \cite{PelikanPazGoldberg:2000}. The Hierarchical problems are thoroughly described in \cite{Pelikan:05}. 

It is also possible to define a noisy version for any of the previous problems. This is done by adding a non-zero Gaussian noise term to the fitness function.

The source code that implements all the problems mentioned in this section can be found in the file  \textit{src/com/EA/Problem.java}.

As mentioned previously, all configuration and parameter setting options of the EA are in the file \textit{EAParameters.txt}. 

To choose a particular problem the user must set the value of the following three options:\\

\begin{tabular}{rl}
~~Line 81:  &  \textit{problemType}\\
~~Line 90:  & \textit{stringSize}\\
~~Line 107: & \textit{sigmaK} ~~~~~(defines the noise component)
\end{tabular}\\

\subsection{How to implement a new problem}

\begin{figure}[h!]
\centering
\includegraphics[width=0.85\textwidth]{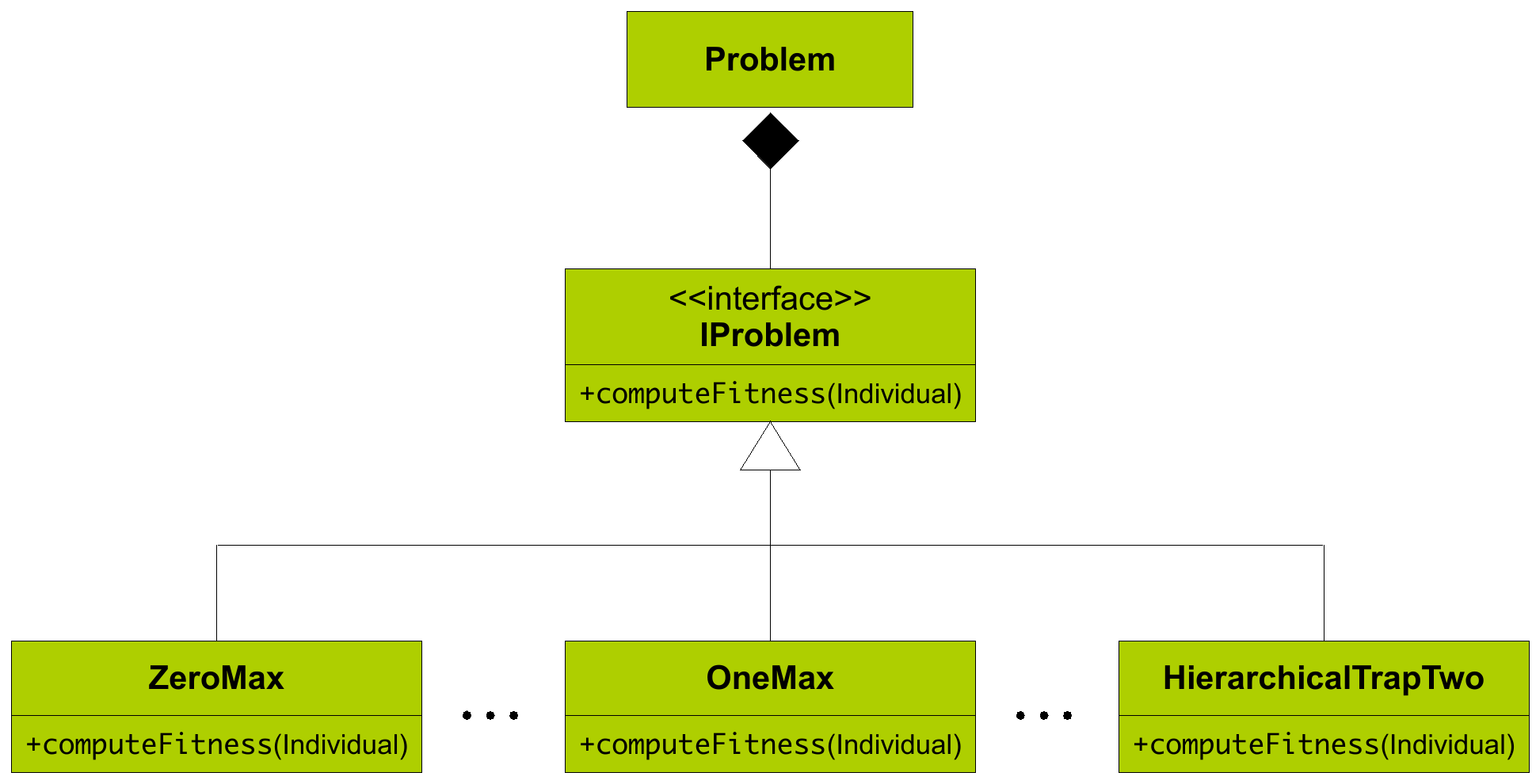}
\caption{The EA uses the design pattern strategy \citep{Gamma:95} to allow an easy implementation of new problems to be solved by the framework.}
\label{fig:IProblem}
\end{figure}

The EA uses the design pattern strategy \citep{Gamma:95} to decouple the implementation of a particular problem from the remaining evolutionary algorithm (see Figure \ref{fig:IProblem}). As a consequence, to integrate a new problem to the given set of test problems it is only necessary to define one class that implements the interface \textbf{IProblem} and change some input options to include the new choice. The interface \textbf{IProblem} can be found in the file \textit{src/com/EA/Problem.java}.  We note that, the reading of these instructions is best complemented with a good analysis of the corresponding source code.

In the following let us consider that we want to solve a new problem called \textit{NewProblem} with the EA algorithm. To plug in this problem it is necessary to:

\begin{enumerate}
\item Define a class called \textit{NewProblem} in the file \textit{src/com/EA/Problem.java}. The signature of the class will be
\begin{figure}[h!]
\centering
\includegraphics[width=0.65\textwidth]{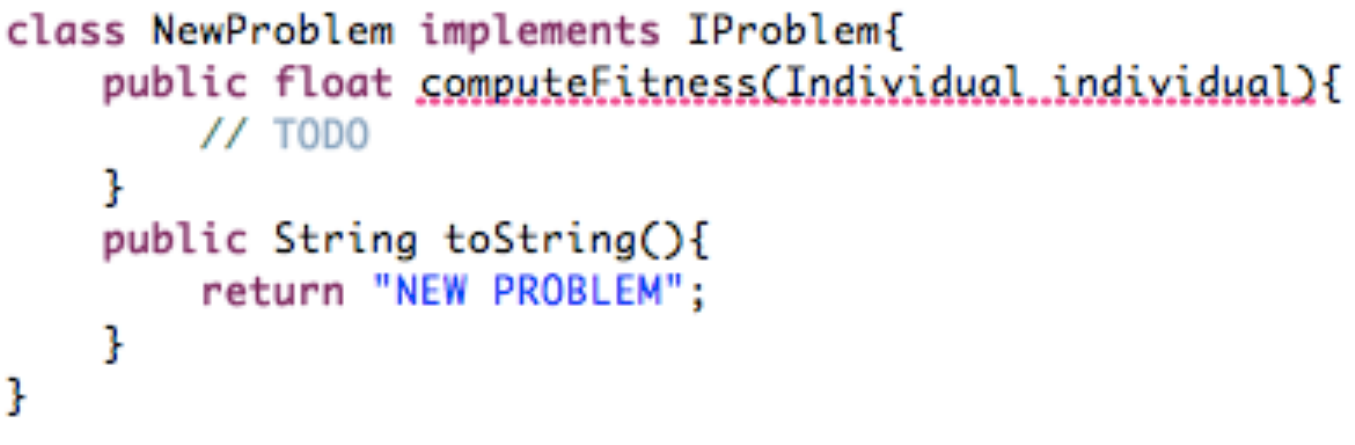}
\end{figure}
\vspace{-.5cm}
\item Code the body of the function computeFitness(Individual) according to the nature of problem \textit{newProblem}. The class \textit{Individual} provides all the necessary functionalities to operate with the string of zeros and ones that represents an individual  (e.g., getAllele(int)). This class can be found in the file \textit{src/com/EA/Individual.java}.

\item To define the new problem option, add the line

\includegraphics[width=.95\textwidth]{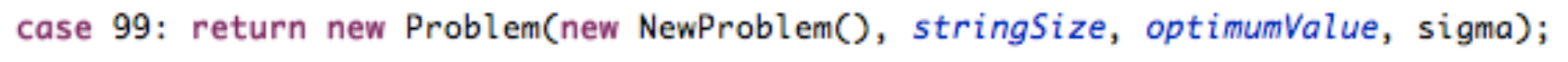}
 to the \textbf{switch} command in the function \textbf{initializeProblem()} of the file \textit{src/com/EA/EAParameter.java}. The case number -- 99 -- is a mere identifier of the new problem option. The user is free to choose other value for this purpose. The rest of the line is to be written verbatim. 
 
\item Validate the new problem option value -- 99 -- by adding the case \textit{problemType} == 99 within the conditional \textbf{if(optionName.equals("problemType"))} of the same \textit{EAParameter.java} file.

\end{enumerate}

Although not strictly necessary, it is also advisable to keep updated the problem menu in the file \textit{ParParameters.txt}.

As mentioned before, the four Java implementations follow the same basic design. In practice, this means that all implementations share some common core Java classes. Here is a brief description of those classes:

\begin{description}
\item[Population] Contains the array of individuals that are the population. It provides all the functionalities necessary for operating that set of individuals. It is also responsible for computing its own statistics such as average fitness and best fitness, which are part of the necessary information to analyze the behaviour of the EAs.

\item[RandomPopulation] Subclass of the class Population. The constructor of this class is responsible for generating the initial random population.   

\item[SelectedSet] Subclass of the class Population. For some algorithms, depending on the selection operator, the size of this set is different from the population size.

\item[Individual] Contains the string of zeros and ones that represents an individual. It provides all the functionalities necessary for operating that string. In particular, it is responsible for computing its own fitness, according to the problem at hand.

\item[Stopper]  Contains all the stop criteria used by the EAs. These criteria are integrated in a single function called \textit{criteria}(...) which in turn must be returned by the \textit{nextGeneration()} function in the file \textit{EASolver.java}. The criteria options can be changed in the file \textit{EAParameters.txt}

\end{description}

\section*{Acknowledgements}
All of the developed source code is one of the by-products of the activities performed by the PhD fellow, José C. Pereira within the doctoral program entitled \textit{Search and Optimization with Adaptive Selection of Evolutionary Algorithms}. The work program is supported by the Portuguese Foundation for Science and Technology with the doctoral scholarship SFRH/BD/78382/2011 and with the research project PTDC/EEI-AUT/2844/2012.



\end{document}